\documentclass[preprint]{acmart}

\AtBeginDocument{%
  \providecommand\BibTeX{{%
    \normalfont B\kern-0.5em{\scshape i\kern-0.25em b}\kern-0.8em\TeX}}}

\setcopyright{none}

\usepackage{amsmath,amssymb,amsfonts}
\usepackage{algorithmic}
\usepackage{graphicx}
\usepackage{textcomp}
\usepackage{comment}
\usepackage{flushend}
\usepackage{subcaption}
\usepackage{gnuplottex}
\usepackage{xcolor}
\def\BibTeX{{\rm B\kern-.05em{\sc i\kern-.025em b}\kern-.08em
    T\kern-.1667em\lower.7ex\hbox{E}\kern-.125emX}}

\begin{document}

\newcommand{\eg}{\textit{e}.\textit{g}., }

\title{TASO: Time and Space Optimization for Memory-Constrained DNN Inference}

\author{Yuan Wen}
\email{weny@tcd.ie}
\affiliation{%
  \institution{Trinity College Dublin}
}

\author{Andrew Anderson}
\email{andersan@tcd.ie}
\affiliation{%
  \institution{Trinity College Dublin}
}

\author{Valentin Radu}
\email{vradu@inf.ed.ac.uk}
\affiliation{%
  \institution{University of Edinburgh}
}

\author{Michael O'Boyle}
\email{mob@inf.ed.ac.uk}
\affiliation{%
 \institution{University of Edinburgh}
}

\author{David Gregg}
\email{david.gregg@cs.tcd.ie}
\affiliation{%
  \institution{Trinity College Dublin}
}



\begin{abstract}
\noindent Convolutional neural networks (CNNs) are used in many embedded applications, from industrial robotics and automation systems to biometric identification on mobile devices. State-of-the-art classification is typically achieved by large networks, which are prohibitively expensive to run on mobile and embedded devices with tightly constrained memory and energy budgets. We propose an approach for ahead-of-time domain-specific optimization of CNN models, based on an integer linear programming (ILP) for selecting primitive operations to implement convolutional layers.
We optimize the trade-off between execution time and memory consumption by: 1) attempting to minimize execution time across the whole network by selecting data layouts and primitive operations to implement each layer; and 2) allocating an appropriate workspace that reflects the upper bound of memory footprint per layer. These two optimization strategies can be used to run any CNN on any platform with a C compiler.
Our evaluation with a range of popular ImageNet neural architectures (GoogleNet, AlexNet, VGG, ResNet and SqueezeNet) on the ARM Cortex-A15 yields speedups of  
8$\times$ compared to a greedy algorithm based primitive selection, reduces memory requirement by 2.2$\times$ while sacrificing only 15\% of inference time compared to a solver that considers inference time only. In addition, our optimization approach exposes a range of optimal points for different configurations across the Pareto frontier of memory and latency trade-off, which can be used under arbitrary system constraints.

\end{abstract}



\keywords{neural  network  optimization, computing operators,  primitive  selection,  optimal  convolutional  layer, memory-time trade off}

\maketitle

\section{Introduction}

CNNs are increasingly deployed in embedded devices for applications such as: vision based perception and control \cite{Levine:2016:ETD:2946645.2946684}, autonomous cars \cite{bojarski16}; in the wearable technology and medical space with wearable cameras assisting visually impaired~\cite{7989772}, to rehabilitate memory loss \cite{sensecam2011}, and to diagnose cancer with human-level efficiency consistently \cite{dermatologist2017nature}.

These emerging computer applications are deployed on mobile and embedded devices, which interact with real world data such as video and audio. With limited compute resources, large neural network models will need to be specially optimized to meet the constraints of each device.
CNNs are not only specific to computer vision, but also find applications in speech recognition \cite{Zhang2016TowardsES} and natural language processing \cite{cnnsentence2014,Kalchbrenner14aconvolutional}. 

However, CNN models often approach hundreds of megabytes (ResNet-50 is about 150 megabytes).
Significant research effort has been invested in reducing model size, with MobileNet~\cite{DBLP:journals/corr/HowardZCKWWAA17} and SqueezeNet~\cite{DBLP:journals/corr/IandolaMAHDK16} being prominent examples of smaller models. However, even reduced-sized models often require too much processing and memory capacity that is available on resource-constrained embedded systems. The alternative is to drastically improve the computational and memory efficiency of larger models to meet the hardware constraints on target platforms -- this is the focus of our work.

In this paper, we address the problem of optimizing the execution of CNNs under tight inference time and memory footprint budgets. Many different implementations exist for performing the operations of convolutional layers, some sacrificing memory for performance (\eg\textit{im2col}), while others applying radical transformations to benefit from cheaper operations (\eg\textit{winograd}). Further, the execution time of an algorithm depends partly upon the layout of data in memory. For example, a common data layout for inputs is $CHW$ (channels, height width), but some algorithms are faster on a $HWC$ data layout.

Using a library of more than 57 primitives implementing convolution via different algorithms and data layouts, we evaluate the optimal primitive selection for each convolutional layer with respect to the overall execution time of the network, and the memory footprint. While other solvers have been used for optimising the primitive selection for inference time ($e.g.$, PBQP~\cite{DBLP:conf/cgo/AndersonG18}), on embedded systems memory constraint is more important and has not been considered for primitive selection before. We explore two optimization strategies: 1) focused on enhancing neural network performance if input, output and working data for all layers of the neural network can fit into main memory; and 2) introducing a \textit{workspace} which is reused by each layer at a time, with the size determined by the ILP optimizer. This ensures that neural networks can be executed on a device with very small memory size, enough to accommodate the requirements of at least on primitive of a convolutional layer, which otherwise would have been impossible to run via a default memory-hungry primitive ($e.g.$ im2col). 

This paper makes the following contributions:

\begin{itemize}

\item We develop an ILP formulation which can expose the trade-off between execution time and memory footprint of CNNs with varying primitives across convolutional layers.

\item We define two optimization strategies for the ILP optimizer, one enhancing the performance across the whole network, while the second imposing a memory budget per layer, which can be applied to whole network in memory and layer by layer in memory respectively.

\item We evaluate our approach using 5 popular CNN (GoogleNet, VGG, AlexNet, SqueezeNet and ResNet) on an edge processor (ARM Cortex-A15).

\end{itemize}

The remainder of this paper proceeds as follows. Motivation for this work is presented in Section~\ref{sec:motivation}. In Section~\ref{sec:background} we give some background on how convolutional layers in CNNs work. Section \ref{sec:methodology} presents our methodology for selecting primitives, including the full ILP formulation, and this is evaluated in Section \ref{sec:experiment}, followed by comparison with other solvers. This paper ends with related work (Section \ref{sec:related_work}) and conclusions (Section \ref{sec:conclusion}).

\section{Motivation}\label{sec:motivation}

\begin{figure}
    \centering
    \includegraphics[width=0.5\linewidth]{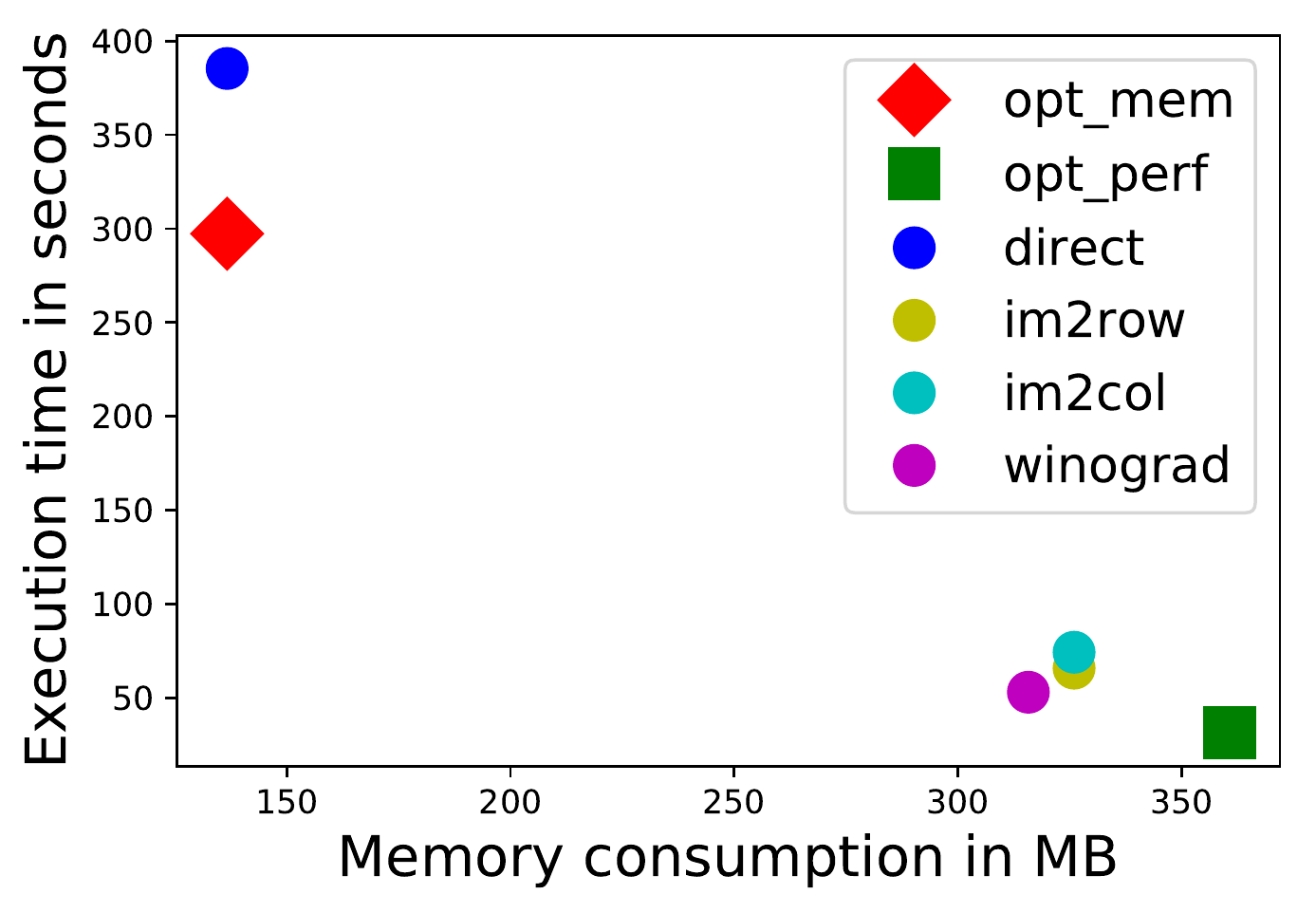}
    \caption{Execution time and memory size of GoogleNet on an ARM A15 processor, with the same primitive choice used across the entire network. We also present the optimal memory (\textit{opt\_mem}) and optimal time performance (\textit{opt\_perf}) with primitives selection from across our full range of 57 primitives.}
    \label{fig:network_sizes}
\end{figure}

Intelligent systems are becoming pervasive at an accelerated rate \cite{Perera2017}, although many of these systems still rely on sending data to the cloud to perform inferences with heavy machine learning models. However, this approach is under increasing scrutiny with pressure to perform more of these computations on device. Sending
data to the cloud requires large amounts of energy for data transmission, and creates privacy concerns for the users of these devices.

\subsection{Intelligence Migrating to Edge Computing Systems}

There is a growing interest to port machine learning to edge computing devices due to a growing suite of applications.

Another reason for migrating computations to the edge is to operate without Internet connectivity even in remote locations, whiles also being independent of connection quality (critical for some applications like self-driving cars). It is thus expected that more deep neural network models will need to be run on devices with tight computation and memory constraints.

\subsection{Running Large Models on Constrained Hardware}

Current deep neural network models are mostly designed for high performance server-side computations where memory constraints are more relaxed. Adopting these models on smaller devices is not straightforward. Figure~\ref{fig:network_sizes} shows the run-time memory requirement of GoogleNet (fitting the whole model into memory), implementing the convolutional layer with different computing primitive choices. One option is to implement the network with a single primitive choice across all convolutional layers (as commonly done by most deep learning frameworks): direct convolution (\textit{direct}), image to row (\textit{im2row}), image to column (\textit{im2col}) and winograd; and the alternative is to select primitives at each convolutional layer to optimize for memory footprint (\textit{opt\_mem}) and for inference time performance~\cite{DBLP:conf/cgo/AndersonG18} (\textit{opt\_perf}). One choice of primitives across the whole network may result in very poor performance (direct convolution running one inference of GoogleNet in 400sec) or may not be able to run the network at all, if physical memory on device is less than 300MB. In contrast, optimizing primitive selection and data layouts for each layer enables more control, not just better execution time (\textit{opt\_perf}) and minimal run-time memory footprint (\textit{opt\_mem}), but also additional optimal points in between to adjust the trade-off for device and application conditions.

Another approach to enable large models on smaller devices is through compression, quantization and pruning of models~\cite{han2015deep_compression}. However, these require extensive effort from data scientists to synthesize new models from original large ones and despite their effort these still achieve lower classification accuracy than original models. To reduce migration cost and to retain the intended inference quality, we consider only methods that produce the same convolutional operation, albeit implemented with different primitives that produce the intended output.

\subsection{Convolutional Layer Workspace}

\begin{figure}
    \centering
    \begin{gnuplot}[terminal=pdf,terminaloptions={font ",9" linewidth 2 size 3.1in,1.3in}]
    
    set boxwidth 0.9
    set style fill solid 
    set datafile separator ","
    set xtics rotate by -45 
    set ylabel "Memory footprint (MB)"
    set logscale y
    set key left top
    set style data histogram
    set style histogram cluster gap 1
    
    plot "graphs/max_memory.csv" every ::1 using 2:xticlabels(1) title "Direct Conv." linetype rgb "#f4a742",\
    "" every ::1 using 3 title "GEMM" lt rgb "#429ef4",\

    \end{gnuplot}
    \vspace{-0.5cm}
    \caption{Memory footprint at run-time of largest layers in some of the most popular deep neural networks.}
    \label{fig:max_memory}
    \vspace{-0.5cm}
\end{figure}

On many devices, memory is very limited (tens of MB) so loading the entire network is not a viable approach. In this situation, computations can be performed layer by layer, loading only the weights necessary for one layer at a time. However, each convolutional primitive has different memory requirements. Figure~\ref{fig:max_memory} shows that one order of magnitude difference exists between the memory requirement of performing the convolution with direct versus using the \textit{im2col} approach. This difference in the amount of memory required by each primitive should instruct the primitive selection to: a) make sure that a layer can fit into memory b) exploit the available space in memory to choose a better performing primitive within the memory budget. This memory budget is what builds our workspace, which is presented in the following sections.

\section{Background} \label{sec:background}

Convolutional neural networks consist of a graph of \textit{layers}. Each layer is a standard
component such as convolution layers, pooling layers, and fully-connected layers. The great
majority of execution time in CNNs is spent in convolution layers \cite{Jia:EECS-2014-93}.
Convolution layers in CNNs operate somewhat differently in traditional convolution in image
processing. Rather than applying a single convolution kernel to a single image, CNNs convolve
an inputs with many channels (typically dozens or hundreds) with many-channel kernels. Further,
in each convolution layer in the input is convolved with many (typically 32-1024) different
multichannel kernels. The result is that convolution layers operate on large input and kernel
tensors, and are computationally very intensive.

There are a number of different algorithms for computing CNN convolution, each of which
has its own strengths and weaknesses. One the earliest successful approaches, \textit{im2col},
lowers the input (image) into a column matrix, and then computes the multiple-channel multiple-kernel
convolution using matrix multiplication. This allows the use of highly-optimized matrix multiplication
libraries, which are available for most processors. However, for a $K \times K$ convolution, lowering
increases the size of the input by a factor of $K^2$. In contrast, the Memory-Efficient Convolution (MEC)
algorithm performs $K$ separate matrix multiplications, and increases the input size by a factor of $K$ \cite{Cho:2017}. The GEMM-accumulating \textit{kern2row} algorithm requires a only a sub-linear increase in space, and performs $K^2$  matrix multiplications \cite{Anderson:2017}. Another approach is to
implement convolution as a simple loop nest, but this loses the benefit of being able to leverage high-performance matrix multiplication libraries.

Fast convolution algorithms rely on mathematical identities to reduce the number of operations
needed. By far the best known fast algorithm is FFT convolution, which reduces the complexity
of convolving two signal of length $n$ from $O(n^2)$ to $O(n log(n))$. However, the dimensions of the convolution
kernels in CNNs are almost always very small, typically $3 \time 3$ or $5 \times 5$, and
algorithms that are specifically designed for short, real-valued convolutions typically
require fewer operations. Winograd convolution \cite{Lavin:2015} greatly reduces the operation
count for CNN convolution, at the cost of linear transforms of the input and output
which require additional computation and significant memory.

CNN frameworks such as TensorFlow, Pytorch and Caffe implement convolution layers by calling
out to specialized libraries that implement these various convolution algorithms.

\section{Primitive Selection with Integer Linear Programming } \label{sec:methodology}

We model the task of configuring a DNN to achieve the best inference time under memory constraint as an integer linear programming (ILP) problem. Typical DNNs can be viewed as a graph of layers, where convolution layers dominate the execution time, and often consume a significant amount of memory. Each convolutional layer in the graph can be implemented by one of a large set of candidate \emph{primitives}, each with their own execution time (\textit{cost}), input and output data transformation costs, and memory requirement. 

In prior work, Partitioned Boolean Quadratic Programming (PBQP) was shown to be efficient in minimizing the execution time of convolutiona neural networks by selecting the optimal primitives for each layer ~\cite{DBLP:conf/cgo/AndersonG18}. However, that earlier work does not model memory requirements, and cannot set a maximum memory budget. This aspect is crucial to embedded devices with limited memory.  Here, we propose a novel approach that optimizes for both memory budget and execution time, which is a much harder problem.  Unlike previous work, our control over memory requirement allows fine adjustments of the time-memory trade-off to generate the optimal network for any application requirements.

\subsection{Integer Linear Programming}
\noindent \textit{Linear programming} is a mathematical model that seeks to minimize an objective function subject to linear constraints. In \textit{Integer} Linear Programming, all decision variables are constrained to integers and hold a linear relationship in the model. ILP solver searches for
a solution that meets the constraints of all the linear relationships,
while seeking to minimize the objective function.
We use the formulation in ~\cite{Papadimitriou:1982:COA:31027} to define this optimisation:

\begin{equation}\label{eq:longeq}
\begin{array}{rrclcl}
\displaystyle \min_{x} & \multicolumn{3}{l}{c^T x} \\
\textrm{s.t.} & A x & \leq & b \\
& x & \geq & 0, & & x \in \mathbb{Z}^{n} \\
\end{array}
\end{equation}

\noindent where 
$x$
represents decision integer variables to be determined, \(\displaystyle c ,b\) are coefficient vectors, \(\displaystyle A\) is a coefficient matrix, and 
\(\displaystyle Ax \leq b\) and \(x \geq 0\) 
are the constraints. 
We search for the solution $x$, an integer sequence which satisfies the given constraints with the lowest cost objective function.

\subsection{ILP Model}

\noindent Equations~\ref{eq:model_exe} and \ref{eq:model_mem} show how our task can be optimised for two objectives, 
for DNN execution time and for memory requirement, respectively. 
In this formulation, decision variable \(\displaystyle x\) is a vector of primitives selected to implement each layer of the DNN. Objective functions \(\displaystyle c_{execution}^{T}x\) and \(\displaystyle c_{memory}^{T}x\) are the \emph{accumulated} execution time and memory usage, respectively. Whereas restrictions \(\displaystyle A_{mem}x \leq M\) and \(\displaystyle A_{exec}x \leq E\) are user-supplied constraints on total execution time ($E$) and memory ($M$) budgets for an inference with the configured DNN.

ILP finds the optimal solutions, one or multiple solutions if these exist, or no acceptable solution if one does not exist. We choose the first determined solution when there are more than one solutions. 
If the is no memory or inference time budget provided (no imposed constraints) then ILP finds the global optimal $x$ under the objective function.
In that situation, ILP searches for the primitive configuration yielding either best performance (using \(\displaystyle c_{execution}^{T}x\), the same as using PBQP~\cite{DBLP:conf/cgo/AndersonG18}) or least memory consumption (using \(\displaystyle c_{memory}^{T}x\)).

\begin{equation}\label{eq:model_exe}
\begin{array}{rrclcl}
\displaystyle \min_{x} & \multicolumn{3}{l}{c_{execution}^{T}x} \\
\textrm{s.t.} & A_{mem} x & \leq & M \\
& x & \geq & 0, & & x \in \mathbb{Z}^{n} \\
\end{array}
\end{equation}

\begin{equation}\label{eq:model_mem}
\begin{array}{rrclcl}
\displaystyle \min_{x} & \multicolumn{3}{l}{c_{memory}^{T}x} \\
\textrm{s.t.} & A_{exec}x & \leq & E \\
& x & \geq & 0, & & x \in \mathbb{Z}^{n} \\
\end{array}
\end{equation}

\subsection{Component Costs}
\noindent The overall costs of a solution comprising total execution time and total memory footprint is composed of several component costs.

The execution time consists of primitive execution time itself and the time spent on data layout transformations of input tensor. Note that we do not use a single canonical data layout for the input, output and kernel tensors used by the convolution primitive. Instead, each primitive specifies the data layouts for input and output, such as $CHW$ or $HWC$ ($C$ being the number of channels, $H$ and $W$ input height and width respectively). These layouts have a different performance on a devices under various convolution algorithms, consuming a specific amounts of memory based on the layer configuration. In addition, if the output format of a chosen primitive does not match the input format of that layer, our framework has to insert a data layout transformation routine to convert the output layout. The execution time and memory cost of these data layout transformations are also considered in our solver. 

Formula~\ref{eq:time_breakdown} presents this execution time for a layer comprised of primitive execution time and data layout transformation time. If the input format is the same as the required output format then data transformation is not needed, associating a zero cost. On the memory side, there is no requirement for extra memory to perform these transformations, since the input and output buffers can be used in place.

The objective functions are expanded in formulas~\ref{eq:time_breakdown}, \ref{eq:algm_exe}, \ref{eq:algm_trans} and \ref{eq:mem} for a network with \(\displaystyle m\) convolutional layers, where \(\displaystyle x_i\) is the selected primitive for the corresponding layer. In formula~\ref{eq:algm_trans}, the matrix \(\displaystyle Trans\_Mat_{i\_(i+1)}\) is a transformation matrix that contains the profiled cost of transforming a tensor data layout between neighbouring convolution layers \(\displaystyle i\) and \(\displaystyle i+1\) given the choice of algorithms selected to implement layers \(\displaystyle i\) and \(\displaystyle i+1\). 

\begin{equation}\label{eq:time_breakdown}
    Time = T\_algm\_exe + T\_format\_trans
\end{equation}

\begin{equation}\label{eq:algm_exe}
    T\_algm\_exe = \sum_{i=1}^{m} x_i \times t_i
\end{equation}

\begin{equation}\label{eq:algm_trans}
\begin{aligned}
T\_format\_trans &= \sum_{i=1}^{m-1}Trans_{i\_(i+1)}[x_{i}, x_{i+1}]
\end{aligned}
\end{equation}

\begin{equation}\label{eq:mem}
    Memory = \sum_{i=1}^{m}x_i \times m_i
\end{equation}

\begin{figure}[ht]
\centering
\includegraphics[width=0.35\textwidth]{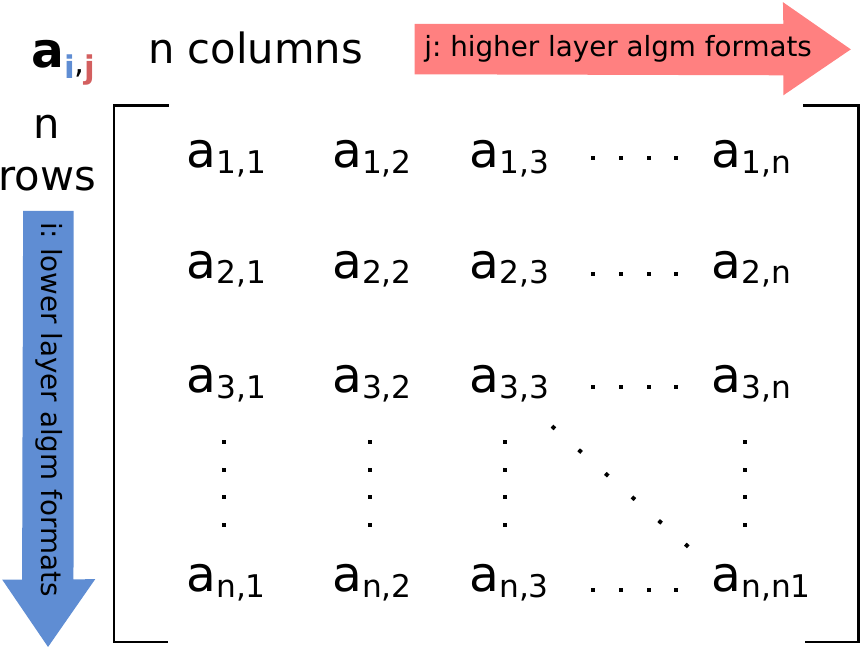}
\caption{Matrix contains the cost for data layout transformation between neighbouring layers.}
\label{fig:trans_matrix}
\end{figure}

Figure~\ref{fig:trans_matrix} shows the data layout transformation matrix. 
Assuming there are \(\displaystyle n \) primitive implementations for both layer \(\displaystyle i\) and following layer \(\displaystyle j\).
The \(\displaystyle Trans\_Mat\) is therefore an n-by-n matrix. 
Each row of the matrix represents the primitive used on layer $i$, and columns present the possible primitives for layer $j$.
Each value on this matrix represents the data layout transformation cost for going from the output format of the first layer to the input format of its following layer. We measure these values during device profiling.

\noindent For any given layer, exactly one primitive can be selected to perform the convolution. 
Therefore, for each layer, the decision variable \(\displaystyle x_i\) is a vector of size \(\displaystyle n\) with each element having a value of either \(\displaystyle 0\) (primitive not selected) or \(\displaystyle 1\) (primitive selected), with $n$ being the number of candidate primitives.  
Since only one primitive will be selected per layer, the vector of decision variables forms a one-hot vector. 



Equations~\ref{eq:algm_exe_T_vector} and \ref{eq:algm_Mem_vector} show the calculation of primitive execution time and memory footprint, where \(\displaystyle \vec{L_i}\) represents the one-hot condition variable vector for layer \(\displaystyle i\), \(\displaystyle \vec{T_i}\) is the profiled execution time vector for all candidate primitives for a layer, and \(\displaystyle \vec{M_i}\) contains the corresponding memory footprint for each primitive.

\begin{equation}\label{eq:algm_exe_T_vector}
    T\_algm\_exe = \sum_{i=1}^{m}\vec{L_i} \bullet \vec{T_i}
\end{equation}

\begin{equation}\label{eq:algm_Mem_vector}
    Memory = \sum_{i=1}^{m} \vec{L_i} \bullet \vec{M_i}
\end{equation}

Similarly, we can model the time required to perform data layout transformations in the same way as above, obtaining equation~\ref{eq:trans_by_vmv}.

\begin{equation}\label{eq:trans_by_vmv}
\begin{aligned}[l]
T\_format\_trans &= \sum_{i=1}^{m - 1}\vec{L_i} \times Trans\_i - (i+1) \times \vec{L_{i+1}}^{T}
\end{aligned}
\end{equation}

\noindent To notice here is that equation~\ref{eq:trans_by_vmv} is not strictly a linear problem due to the multiplication between every two nodes. However, the decision variable vectors are one-hot vectors, which allows us to transform this equation back to a linear form by evaluating the \emph{floor} function of the sum of the decision variables.

\begin{equation}
    \begin{aligned}
        &Trans_(i, (i+1)) = 
        \sum_{j=1}^{n}\sum_{k=1}^{n} \lfloor (L_{i_j} + L_{(i+1)_k}) / 2 \rfloor \times Trans(i, (i+1))
    \end{aligned}
\end{equation}

\begin{equation}\label{eq:expanded_exeT}
    \begin{aligned}
    Time = &\sum_{n=1}^{m}\vec{L_n}\bullet\vec{T_n} \quad + \quad \sum_{i=1}^{m-1}\sum_{j=2}^{m}\vec{L_i} \times Trans\_(i,j)\times\vec{L_j}^{T}
    \end{aligned}
\end{equation}

\subsection{Whole network performance optimization with ILP} \label{sec:final-model}

This optimization strategy is designed for devices with enough physical memory to accommodate the entire network into main memory, optimizing for performance across the whole network. A memory budget can be introduced if sharing the system memory with other tasks and selecting performance critical conditions to be met. The ILP optimiser finds the optimal primitive selection on each layer of a network to meet the imposed constraints.

\begin{equation}\label{eq:final_model_performance}
\begin{aligned}
minimize \\
\quad Time = &\sum_{i=1}^{m}\vec{L_i}\bullet\vec{T_i} 
\quad + \quad
\sum_{i=1}^{m-1}\sum_{j=2}^{m}\vec{L_i} \times Trans\_(i,j)\times\vec{L_j}^{T} \\
\textrm{s.t.} \quad\quad &\sum_{i=1}^{m}\vec{L_i}\bullet\vec{M_i} \quad \leq Memory Budget \\
&\vec{L_i} =  \{x_j| j \in i \times n \: to \: (i+1) \times n  \: 
\quad \wedge \quad 
n = num \: of \: primitives \: 
\quad \wedge \quad
x_j = 0, 1\} \\
&\sum_{j=1}^{n}\vec{L_{i_j}} = 1, \quad \forall i \in [1,m]\\
\end{aligned}
\end{equation}

Formula~\ref{eq:final_model_performance} optimizes for inference time under constrained imposed by the allowed memory consumption, while Formula \ref{eq:final_model_memory} optimizes memory footprint with constrains from execution time. Varying these constrains allows us to find a Pareto of optimal points under imposed memory or performance constraints as presented in the evaluation section. 

\begin{equation}\label{eq:final_model_memory}
\begin{aligned}
minimize \\
\quad Memory = &\sum_{i=1}^{m}\vec{L_i}\bullet\vec{M_i} \\
\textrm{s.t.} \quad\quad & \sum_{i=1}^{m}\vec{L_i}\bullet\vec{T_i} 
\quad + \quad
\sum_{i=1}^{m-1}\sum_{j=2}^{m}\vec{L_i} \times Trans\_(i,j)\times\vec{L_j}^{T} 
\quad
\leq Performance Budget\\
&\vec{L_i} =  \{x_j| j \in i \times n \: to \: (i+1) \times n  
\quad \wedge \quad
n = num \: of \: primitives \:  \quad \wedge \quad
x_j = 0, 1\} \\
&\sum_{j=1}^{n}\vec{L_{i_j}} = 1, \quad \forall i \in [1,m]\\
\end{aligned}
\end{equation}

\subsection{Workspace-based optimization (layer by layer)}

\begin{figure*}
    \centering
    \includegraphics[width=0.75\linewidth]{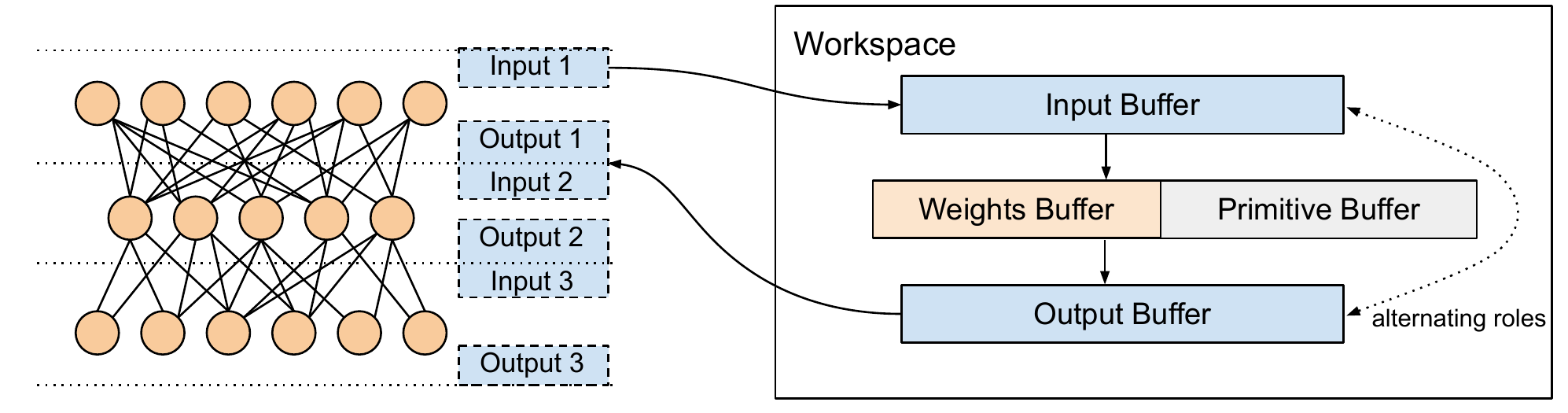}
    \caption{Reusing three allocated buffers (input, output and weights) by flipping the roles of the input and output buffer after each convolutional layer, with the output of one layer becoming the input of following layer.}
    \label{fig:workspace_diag}
\end{figure*}

If device memory is limited such that the entire neural network does not fit into main memory with any combination of primitives, we consider the alternative optimization strategy, layer by layer. This is solved using the ILP solver also. A workspace size is identified to represent the upper bound of memory footprint to be used by primitives for each layer in the network, one at a time. 

Figure~\ref{fig:workspace_diag} presents the component buffers of the workspace. This includes two alternating input-output buffers, a buffer to hold the layer weights. The weights of each layer of the network are loaded into the workspace in progression. Additional buffers hold transformations and intermediary values.

The size of this workspace is determined by equation~\ref{eq:final_model_workspace} below.

\begin{equation}\label{eq:final_model_workspace}
\begin{aligned}
minimize \\
\quad | Workspace |  = &\max_{i=1}^{m}\vec{L_i}\bullet\vec{M_i} \\
\textrm{s.t.} \quad\quad & \sum_{i=1}^{m}\vec{L_i}\bullet\vec{T_i} \quad + \quad \sum_{i=1}^{m-1}\sum_{j=2}^{m}\vec{L_i} \times Trans\_(i,j)\times\vec{L_j}^{T}  \quad \leq Performance Budget\\
&\vec{L_i} =  \{x_j| j \in i \times n \: to \: (i+1) \times n  \: \quad\wedge\quad n = num \: of \: primitives \: \quad\wedge\quad x_j = 0, 1\} \\
&\sum_{j=1}^{n}\vec{L_{i_j}} = 1, \quad \forall i \in [1,m]\\
\end{aligned}
\end{equation}

This method follows a first stage of performance optimization across the whole network as indicated by equation~\ref{eq:final_model_performance} to identify the appropriate workspace size. Equation~\ref{eq:final_model_workspace} differs from equation~\ref{eq:final_model_memory} by performing an optimization over the maximum memory required across all layers, as opposed to accumulating memory when all network layers are simultaneously in memory.
\section{Evaluation} \label{sec:experiment}
To evaluate our approach we implemented the ILP solver and evaluate it in this section on selecting
primitive routines for convolutional layers to run on a low power embedded processor core.
We measure the execution time and memory requirement for each convolutional layer implemented with one of the 57 convolution primitive routines in the profiling stage. These device profile measurements are the input for the ILP solver, which selects the optimal primitive for each layer. 

\subsection{Experimental Setup}
We use the ODroid-XU4 embedded development board, which includes
the Samsung Exynos-5422 processor. The Exynos-5422 has the low-power embedded
core  
2.1GHz ARM Cortex-A15.  
The Cortex-A15 is a powerful 32-bit out-of-order superscalar core, with NEON instructions. We concentrate on the CPU because this is the most common computation unit found in embedded devices, although where available, the GPU can be utilized similarly.

Our ILP solver based approach is not tied to any specific CNN architecture or hardware platform, being generic to determine the best primitive selection solution for any CNN and environment. This is implemented with the open-source PuLP python package.

This evaluation is performed on five popular deep learning models (AlexNet, VGG, GoogLeNet, ResNet and SqueezeNet) due to: (1) they are highly
accurate for image recognition tasks, as shown by their success in the ILSVRC 2014 competition;
(2) they have a relatively complex network
structure, with forks and joins, rather than a simple sequence of layers. 

Our results present only the inference process (not the training phase) of an already trained neural network. The chosen evaluation scenario reassembles a streaming application, processing a single frames at a time.

\subsection{ILP and Constraints}

\begin{figure*}
\begin{subfigure}{0.5\linewidth}
\centering
\includegraphics[width=\textwidth]{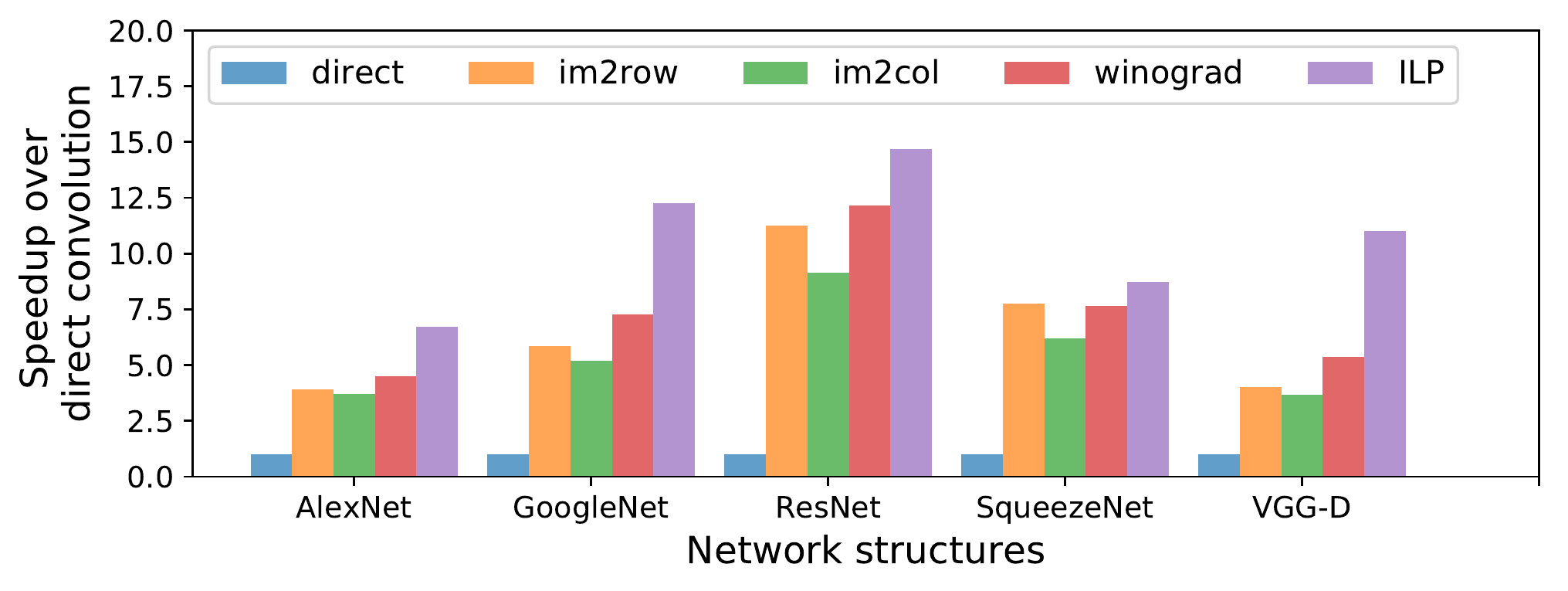}
\caption{Performance speedup over all direct convolution.}
\label{fig:memory_performance_a}
\end{subfigure}%
\begin{subfigure}{0.5\linewidth}
\centering
\includegraphics[width=\textwidth]{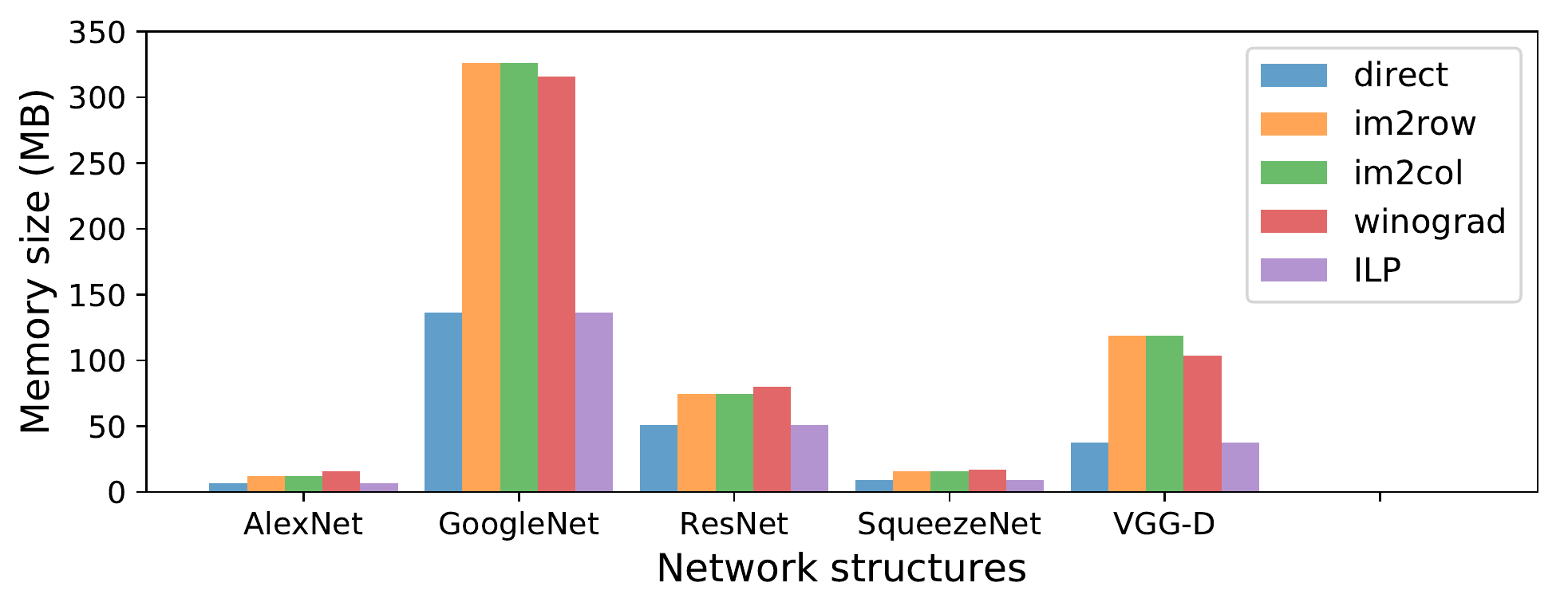}
\caption{Memory footprint.}
\label{fig:memory_performance_b}
\end{subfigure}
\caption{Performance and memory optimisations with ILP for five popular CNNs on the ARM Cortex 15.}
\label{fig:memory_performance}
\end{figure*}

The task in to select the optimal primitive routine for each convolutional layer within a CNN. This is done using the profiled execution time of each primitives on the target device (ODroid-XU4). We optimise the selection with our ILP solver, with the constraint of a given memory budget and optimising the execution time. This generates a Pareto optimal frontier of best primitive selection solutions for varying memory budget. In this experiment, we optimize for best performance over the entire network (whole network in memory).

To find the optimal solution, the ILP solver takes 23.84 seconds on the layers of GoogleNet. 
Table~\ref{tab:ilp_solver_cost} presents such cost for all networks we discussed in this paper. 
Given that configured networks generally work on devices for a long period of time (millions or billions of inferences), the one-time cost of the ahead-of-time optimization can be considered trivial.

\begin{table}[]
    \centering
    \caption{The time required by our solver to find a solution for each CNN, using an Intel Core i5 CPU. This is after each layer has been profiled on the target device in the offline phase.}
    \begin{tabular}{c|c|c}
        Network & Number of conditional variables  &Time (in sec)\\
        \hline
        AlexNet & 456 & 3.373 \\
        VGG & 741 & 5.259 \\
        SqueezeNet & 1710  & 8.835 \\ 
        GoogleNet & 3990 & 23.84 \\
    \end{tabular}

    \label{tab:ilp_solver_cost}
\end{table}

Figure \ref{fig:memory_performance_a} shows the execution time performance for running the five networks implemented with primitives statically allocated in the first four bars (the same primitive used across the whole network) and the optimized primitive selection with ILP on the ARM processor architectures Cortex-A15 with the right-most bar. The \textit{direct}
method is a simple optimized nested loops that implements a convolutional layer. \textit{Im2row}
and \textit{im2col} expand the input tensor and use highly-optimized
matrix multiplication libraries to perform all the convolution operations in one matrix multiplication operation, as previously described in Section \ref{sec:background}.
\textit{Winograd} performs fast convolution by converting to the Fourier domain. Finally,
our \textit{ILP} solver finds the fastest combination of primitives across layers from the list of 57 
primitive implementations (Figure~\ref{fig:memory_performance_a}). These results show 
that the \textit{im2row} method is very fast, but the combination of methods selected by
our ILP method is significantly faster. Since there is no memory constrain in this experiment, the performance of our ILP is the same as that of PBQP~\cite{DBLP:conf/cgo/AndersonG18}, which we confirm.

\begin{figure*}
\begin{subfigure}{0.45\linewidth}
\centering
\includegraphics[width=\textwidth]{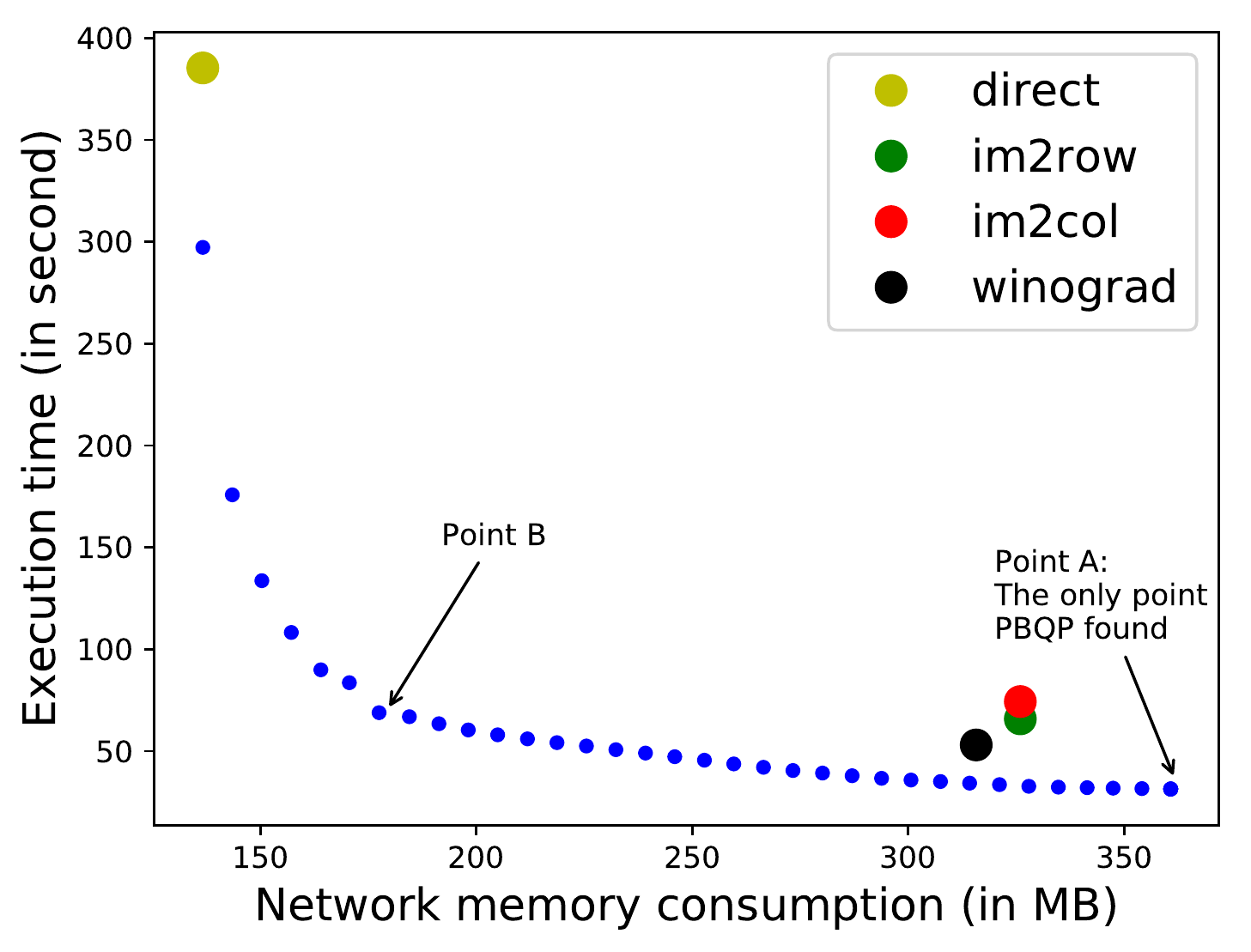}
\caption{Performance optimization across the whole network.}
\label{fig:time_mem_googlenet_a15_2}
\end{subfigure}%
\hspace{0.7cm}
\begin{subfigure}{0.45\linewidth}
\centering
\includegraphics[width=\textwidth]{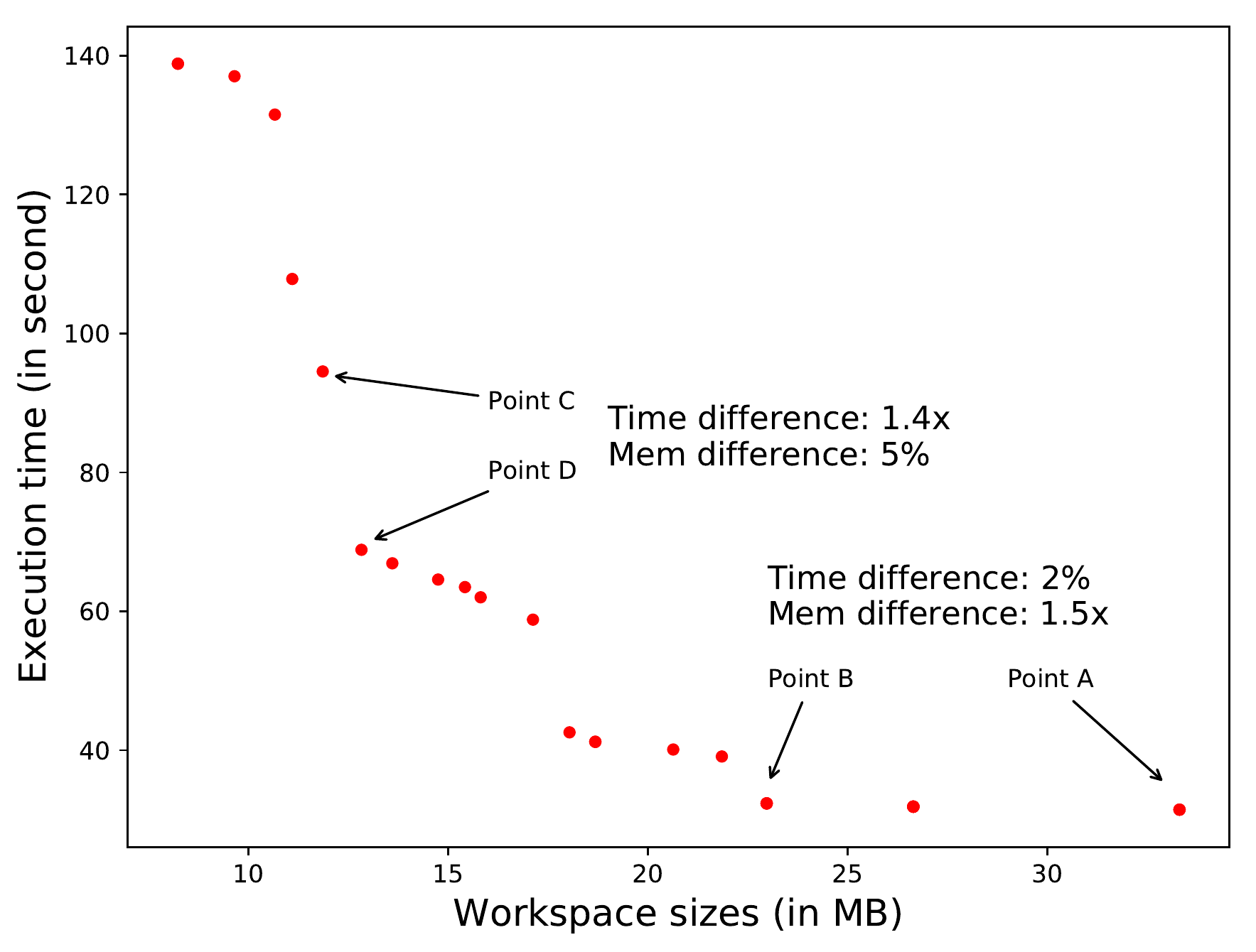}
\caption{Workspace-based optimization with layer level performance.}
\label{fig:time_workspace_googlenet_a15}
\end{subfigure}
\caption{Run-time performance of the two optimization strategies applied to GoogleNet (ImageNet) on the ARM A15 processor.}
\label{fig:time_mem}
\end{figure*}

Unlike PBQP, our ILP solver can also optimise primitive selection to achieve the best memory footprint (Figure~\ref{fig:memory_performance_b}), which is important on devices with tight memory space. Optimising in the memory footprint direction, our ILP solver finds a combination with the smallest memory footprint. This combination is similar in memory requirements to the \textit{direct} method, which theoretically has the smallest memory footprint, although much faster in inference time, as we will seen in the following section. These two experiments demonstrate the strength of our ILP solver to balance constraints and achieve the optimal points when either memory or inference time are critical for the system.

Figure \ref{fig:time_mem} shows the Pareto frontier of the ILP optimal solutions for primitive selection, achieved when varying the memory budget as given condition to the solver. 
Figure~\ref{fig:time_mem_googlenet_a15_2} shows the inference time on a single image against the memory requirement optimised for performance across the whole network.
The solutions to the left of the chart optimize for memory in detriment to execution time. Each blue dot is an optimal solution determined by IPL. We can see that the most memory efficient implementation across the whole network is achieved by the direct convolution, in detriment of inference time. Alternatively, there are many other configuration when memory constraint is relaxed. ILP determines optimal configuration points that are better in inference time than im2col, im2row and even Winograd primitives across the network, despite executing in the same memory budget. The bottom right corner is optimising purely for inference time, with almost no constraint from memory budget (the only point found by the PBQP optimizer~\cite{DBLP:conf/cgo/AndersonG18}). This shows that solutions with different primitive configurations that surpass the performance of only im2col (generally regarded as the faster implementation option) do exist. Another important observation on this Pareto curve is that memory requirement can be reduced dramatically with minimal impact on inference time (more than two thirds of memory requirement can be reduced for just less than half of second inference slowdown). These observations cannot be captured by previous optimization solutions and have significant impact. Our method exposes better combinations of primitives capable of running at the same speed as im2col, but at half the memory requirement of im2col.

\begin{figure}
\centering
\begin{subfigure}{0.45\linewidth}
\centering
\includegraphics[width=\textwidth]{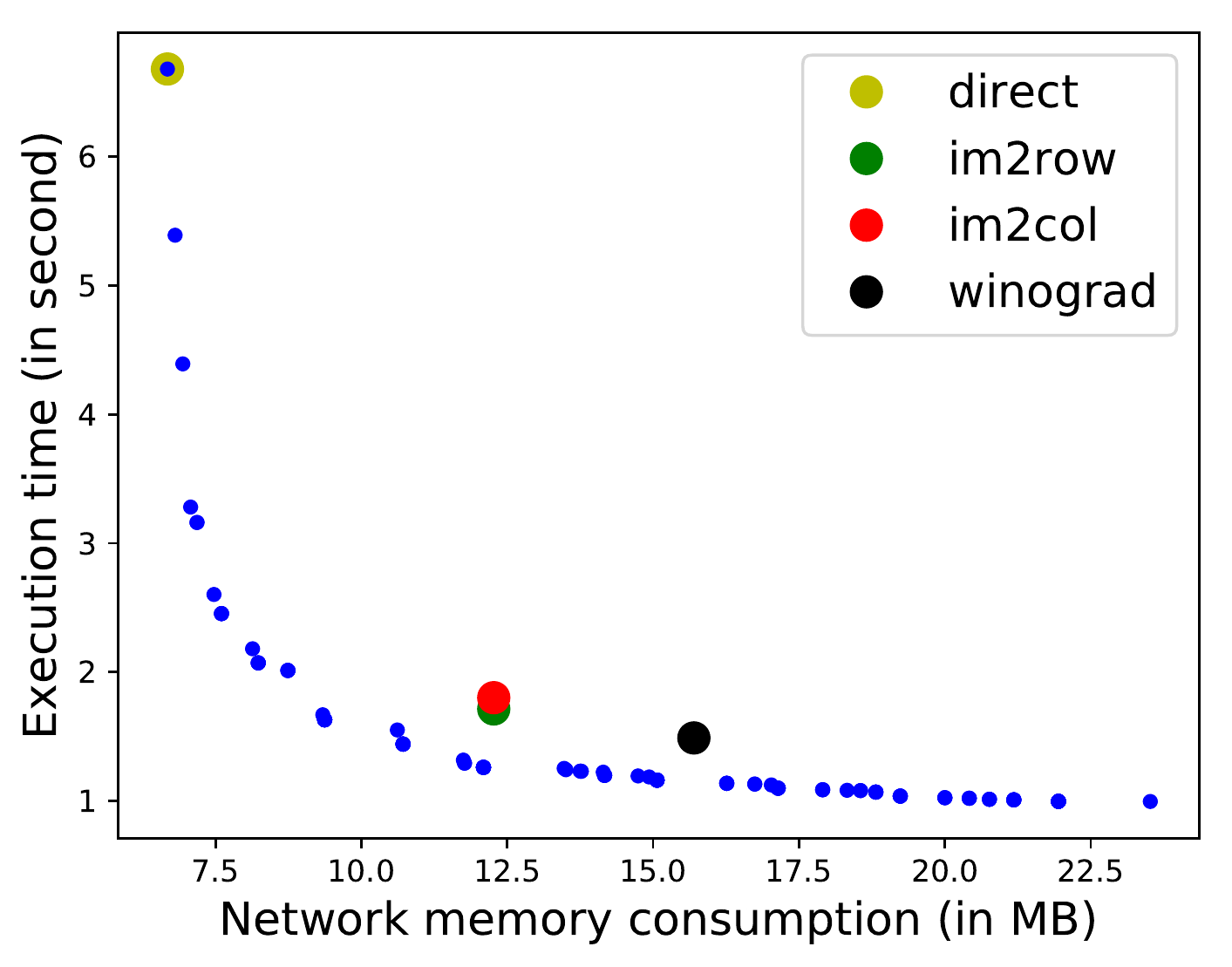}
\caption{Across the whole network.}
\label{fig:time_mem_alexnet_a15_2}
\end{subfigure}
\hspace{0.7cm}
\begin{subfigure}{0.45\linewidth}
\centering
\includegraphics[width=\textwidth]{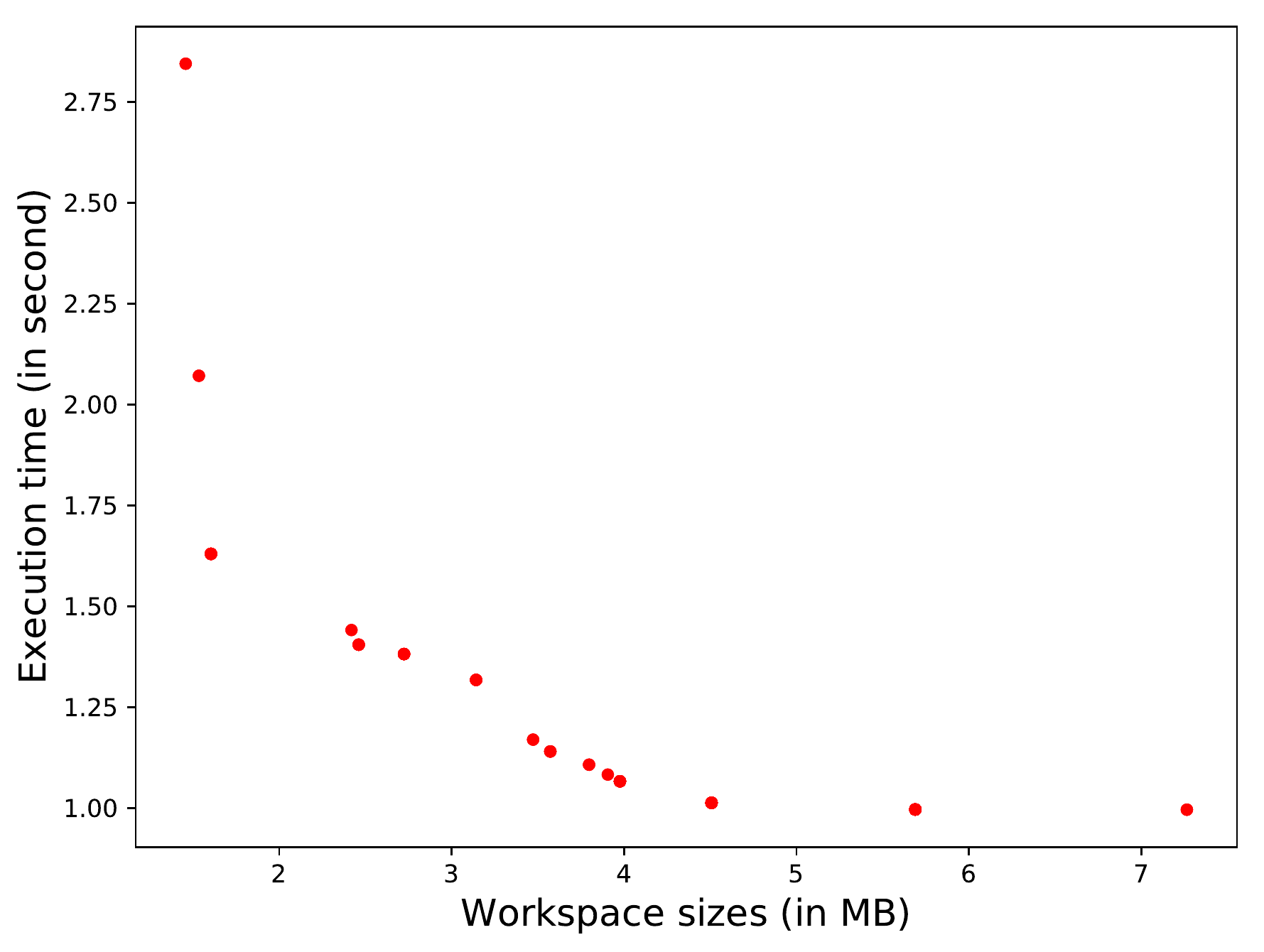}
\caption{Workspace optimization.}
\label{fig:time_workspace_alexenet_a15}
\end{subfigure}
\caption{Run-time performance of the two optimization strategies applied to AlexNet (ImageNet) on the ARM A15 processor.}
\label{fig:alexnet_time_mem}
\end{figure}

\begin{figure}
\begin{subfigure}{0.45\linewidth}
\centering
\includegraphics[width=\textwidth]{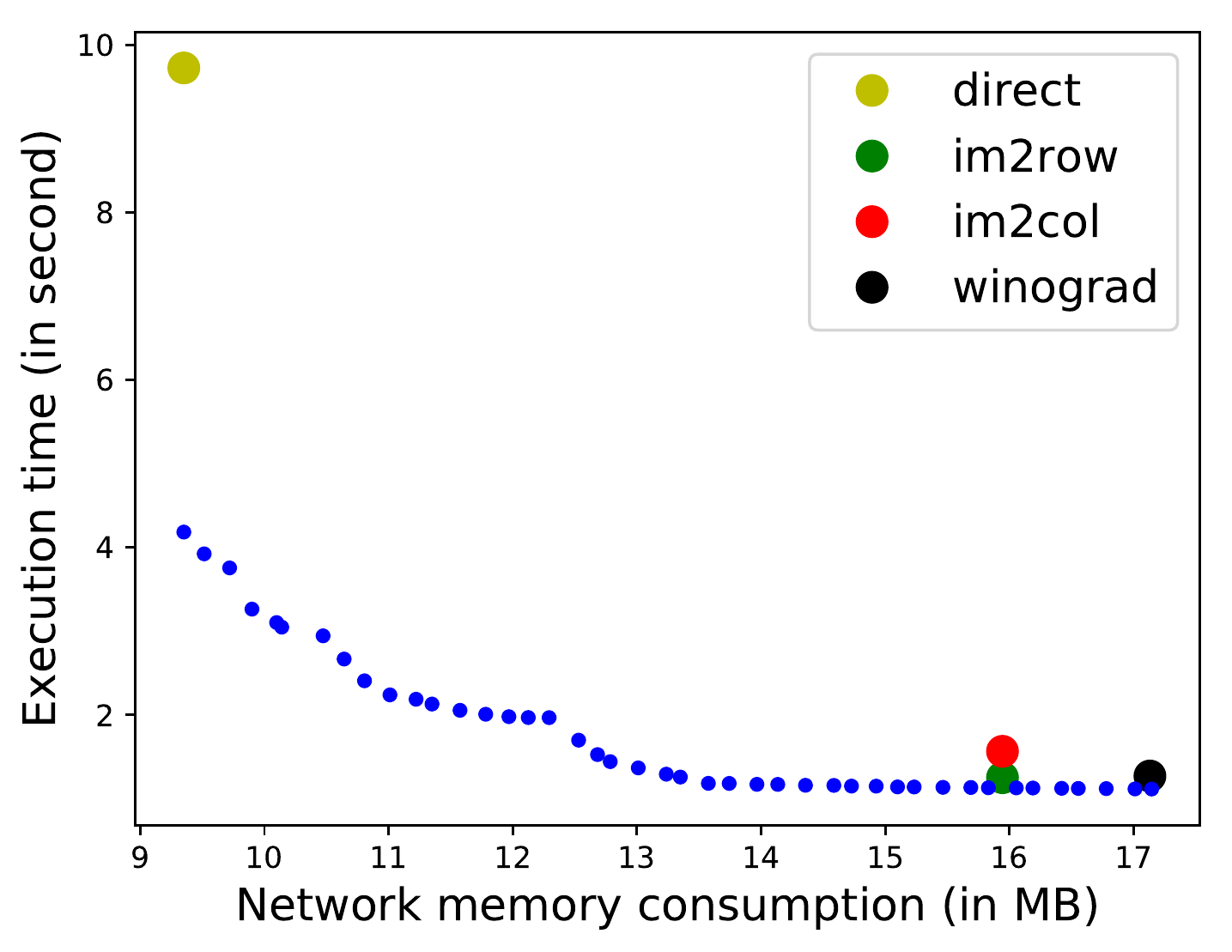}
\caption{Across the whole network.}
\label{fig:time_mem_squeezenet_a15_2}
\end{subfigure}
\hspace{0.7cm}
\begin{subfigure}{0.45\linewidth}
\centering
\includegraphics[width=\textwidth]{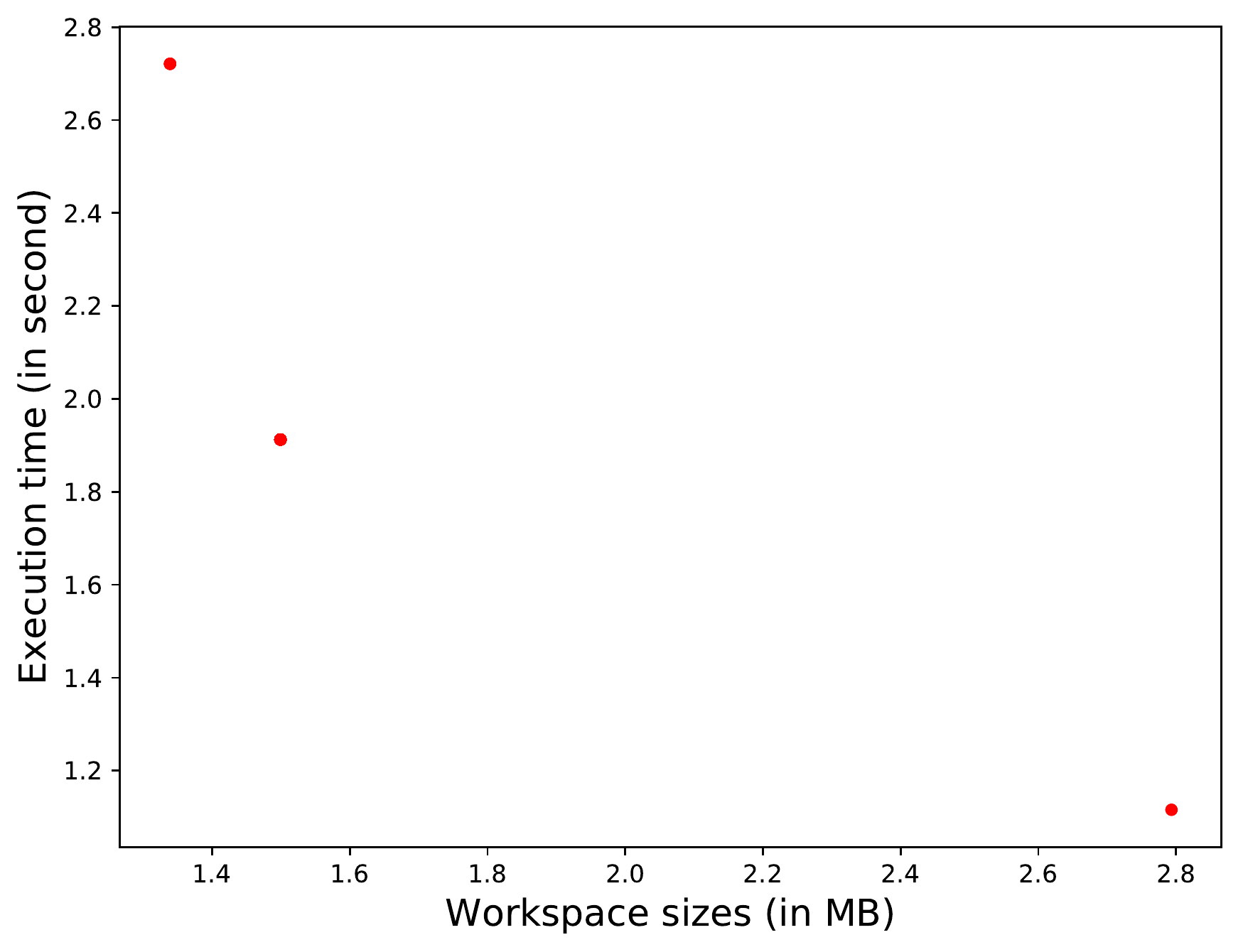}
\caption{Workspace optimization.}
\label{fig:time_workspace_squeezenet_a15}
\end{subfigure}
\caption{Run-time performance of the two optimization strategies applied to SqueezeNet (ImageNet) on the ARM A15 processor.}
\label{fig:squeezenet_time_mem}
\end{figure}

\begin{figure}
\begin{subfigure}{0.45\linewidth}
\centering
\includegraphics[width=\textwidth]{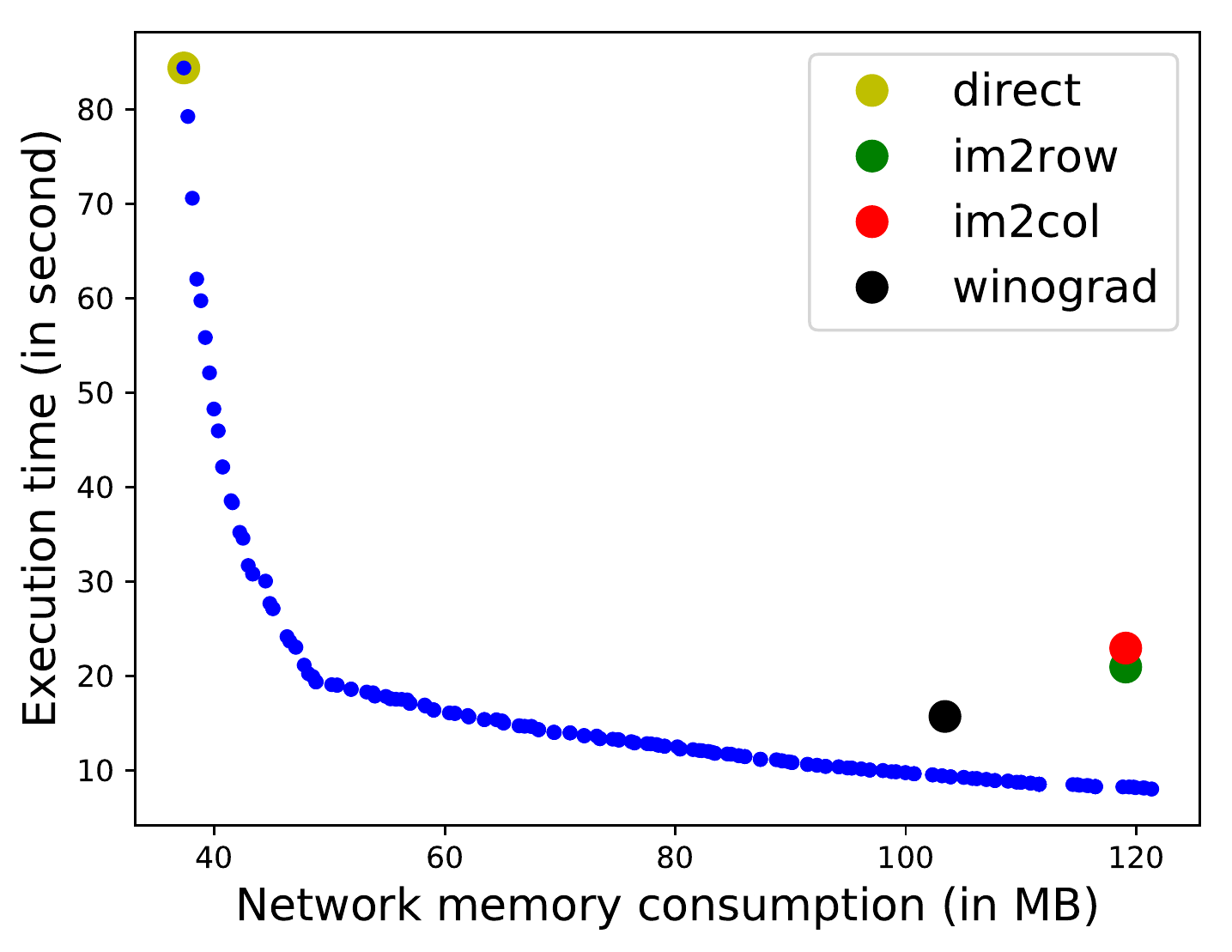}
\caption{Across the whole network.}
\label{fig:time_mem_vgg_a15_2}
\end{subfigure}
\hspace{0.7cm}
\begin{subfigure}{0.45\linewidth}
\centering
\includegraphics[width=\textwidth]{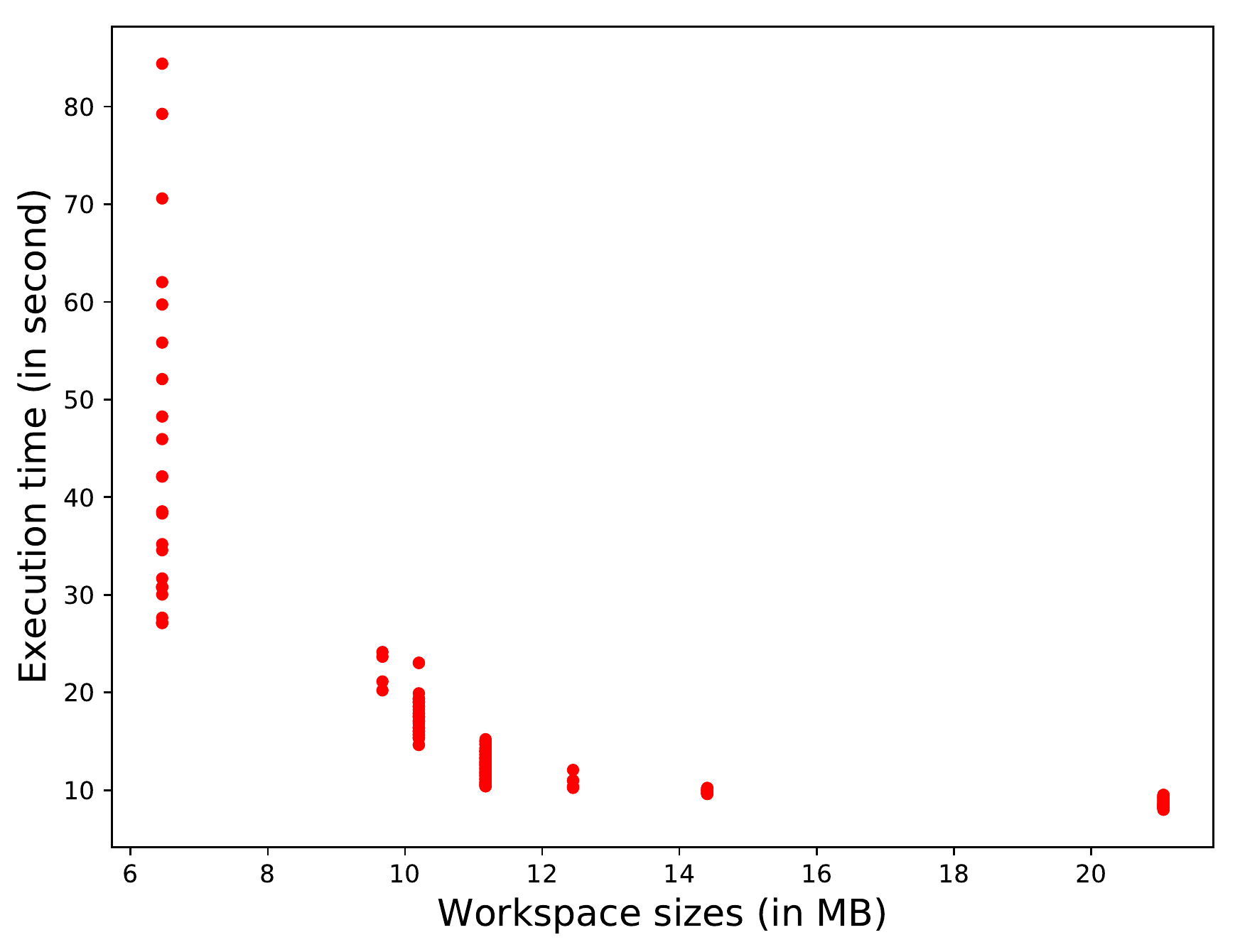}
\caption{Workspace optimization.}
\label{fig:time_workspace_vgg_a15}
\end{subfigure}
\caption{Run-time performance of the two optimization strategies applied to VGG-D (ImageNet) on the ARM A15 processor.}
\label{fig:vgg_time_mem}
\end{figure}

Figure~\ref{fig:time_workspace_googlenet_a15} shows the performance of the other optimization strategy when the whole network does not fit into main memory, but with optimizations applied at layer level using the workspace memory budget strategy. This shows that primitives chosen at point A and point B (as highlighted on the figure) can execute with almost the same latency, but point B uses about 10MB less  memory. Optimising for a tight memory budget (around 12MB) shows that different configurations exist that can achieve a difference of 1.4$\times$ speedup (points C and D on the figure). These observations show that even selections at layer level can have a dramatic impact on overall performance and ILP helps to expose these conditions through the Pareto trade-off curve.

We observe similar patterns in the ILP identified optimal execution points for the other deep neural networks, AlexNet (Figure~\ref{fig:alexnet_time_mem}), SqueezeNet (Figure~\ref{fig:squeezenet_time_mem}), VGG (Figure~\ref{fig:vgg_time_mem}) and ResNet, and across other processors architectures -- similar patterns on the ARM Cortex A7, which we leave out in the interest of space.
Our results demonstrate that our proposed ILP based primitive selection optimiser is network and hardware agnostic, identifying the best solutions under given constraints and optimisation objective (inference time, memory footprint) on any device. 

\subsection{Comparison with Other Methods}

Here we compare our ILP-based solver with other methods for convolutional layer primitive selection. We show that our ILP method offers more control than PBQP in memory constraint conditions and finds better solutions than greedy algorithms.

\subsubsection{PBQP Solver}

Figure~\ref{fig:time_mem_googlenet_a15_2} shows the solutions found by ILP and PBQP~\cite{DBLP:conf/cgo/AndersonG18}. All the red points are optimal solutions provided by ILP under varying execution-time and memory budget constraints, while PBQP solution is only point A, which is optimal for execution-time though completely disregarding memory footprint. Our ILP solver allows for a small sacrifice in inference time, 15\%, to make a huge saving in memory footprint, 2.2$\times$ (point B). The difference between the two solvers is clear, PBQP can find only the fastest combination of primitives, but has no control over the memory footprint, which can be a crucial constraint on embedded devices.

\subsubsection{Greedy Solver}

\begin{figure}[ht]
\centering
\includegraphics[width=0.45\textwidth]{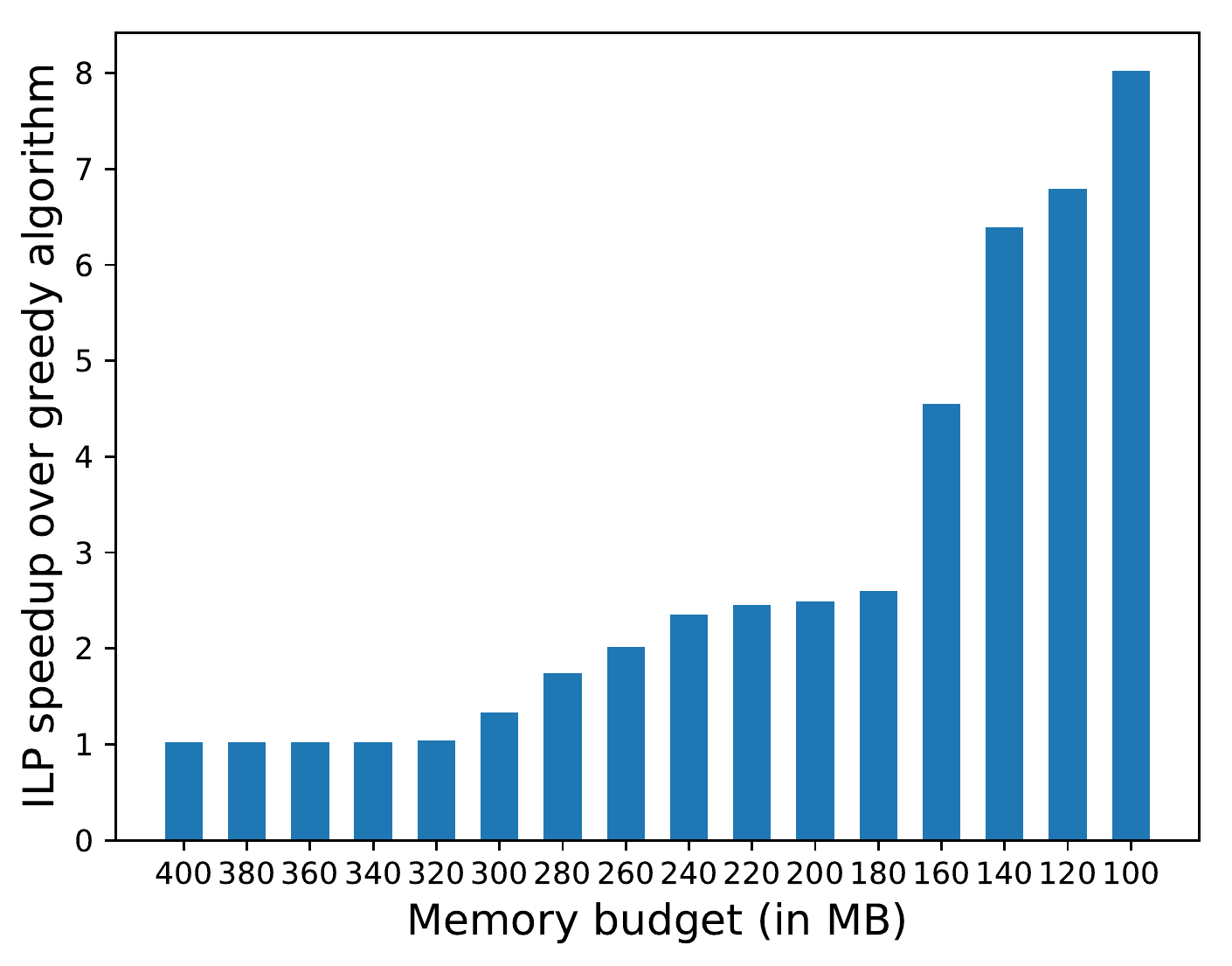}
\caption{Speedup from ILP versus the greedy primitive selection for GoogleNet. Moving left to right on the graphs the memory budget is \textbf{decreased}. Each bar is the speedup of the ILP optimal network arrangement for the given memory budget versus the arrangement found by the greedy heuristic selection. As the memory budget is constrained further, the higher-quality solutions found by ILP quickly outpace the solutions found by greedy heuristic selection.}
\label{fig:versus_A15}
\end{figure}

We implemented a greedy algorithm to consider to add the memory footprint constraint to PBQP in a simple strategy. It first tries to optimize the choice of primitives to obtain the shortest inference time (same as PBQP). After the first selection of primitives per layer, if the memory footprint budget imposed to the optimizer is exceeded, the primitive with the largest memory footprint is replaced by another primitive with smaller memory footprint but fastest among the other remaining candidates, so that inference time is not compromised. This process is repeated by replacing candidate primitives until the constraint of memory budget is achieved.

In Figure \ref{fig:versus_A15} we compare the greedy algorithm with our ILP solver by presenting only the speedup achieved by our method against the greedy algorithm on varying memory budget constraint. This primitive selection is performed on the layers of GoogLeNet across the whole network, running on the ARM Cortex-A15 processor. We can see that while the memory budget is plentiful, both solvers make similarly good decisions (leftmost bars), while with the tightening of memory constrains (rightmost bars) the greedy algorithm begins to provide sub-optimal solutions. This is because there are just a small set of very good solutions when the memory budget is small, whereas when the memory budget increases, the greedy algorithm starts making short-sighted decisions with the simple primitive replacement strategy. In contrast, these solutions are easily identified by our ILP solver, reaching 8x speedups on the processor.

\section{Related Work} \label{sec:related_work}

Several tools for producing optimized code for deep neural networks 
have been proposed in recent years. These tools operate at different 
levels of granularity, from controlling code generation for individual 
neurons right up to run-time heuristics for selecting full layer implementations.

At the fine-grained end of the spectrum is Latte \cite{Truong:2016}, a 
domain-specific language, compiler and run-time for DNNs. Latte programs 
are written using abstractions of neurons, ensembles, and connections that 
allow the programmer to write a very detailed specification of the layers of 
a DNN. 

A sophisticated compiler then produces optimized
code, and pattern-matches for code regions that can be replaced by
calls to optimized libraries such as GEMM. However, like most compilers, 
Latte can only generate code that corresponds to the program which the 
programmer wrote down. Our proposal, and the proposal of the prior work of 
Anderson et al.~\cite{DBLP:conf/cgo/AndersonG18} is to allow the tool to 
have control of algorithmic choice for layer implementations so this burden 
is removed from the programmer.

Compilers for algorithmic choice have been a topic of study; one of the most
widely-known general-purpose examples is the PetaBricks language and compiler from MIT~\cite{DBLP:conf/pldi/AnselCWOZEA09}.

More recent developments include Boda~\cite{Moskewicz:2017}, a program
generator for deep neural networks. Boda generates large numbers of
variants of loop nests to implement DNN layers, and uses auto-tuning
to select from among them. There are two significant advantages to using our
approach over auto-tuning: auto-tuning takes much more time than solving a small 
ILP problem, and can provide at best a heuristic solution. In contrast, our
analytic approach can find the global optimal network instantiation with respect
to the given time and memory constraints.

While Latte and Boda can provide speedups, the majority of DNNs used in practice are still implemented 
using vendor-supplied programming frameworks. A common approach taken in these frameworks is to provide a means 
to query a heuristic cost model embedded in the library at run-time, to select
the method to implement the layer which is about to be executed. This approach is taken 
by NVIDIA's CUDNN framework~\cite{CUDNN:2017}, among others.

In contrast, our approach is to solve for the optimal layer selection ahead of time. 
We profile layer execution to derive costs rather than relying on a heuristic cost model. 

The TensorFlow XLA ahead-of-time compiler~\cite{Dean:2017} is an optimization tool for 
DNNs written in the TensorFlow domain-specific language. XLA performs ahead-of-time
optimizations on the network graph. The techniques we present 
seem well-suited to systems such as XLA, given the inherently ahead-of-time nature
of our proposed approach.

\section{Conclusion} \label{sec:conclusion}

We address a growing challenge: how to make the large deep neural network models available to use on mobile and embedded devices with limited resources. We propose an integer linear programming-based domain-specific optimizer. By selecting algorithms, implementations, and data layouts for each convolutional layer, 
we can optimize for both inference latency and memory footprint. Our results demonstrate how this optimizer works on five popular convolutional neural networks, observing very good performance under both constrains from inference time and memory footprint budgets when compared with greedy algorithms and performance-only solvers for convolutional layer primitive selection. We believe this opens the opportunity for many innovative edge application to build on the higher accuracy achievable only by large deep neural network models which have been unpractical on these small devices until now.


\bibliographystyle{ACM-Reference-Format}
\bibliography{egbib}

\end{document}